\documentclass{vgtc}                          

\ifpdf
  \pdfoutput=1\relax                   
  \usepackage{graphicx}                
  \DeclareGraphicsExtensions{.pdf,.png,.jpg,.jpeg} 
\else
  \usepackage{graphicx}                
  \DeclareGraphicsExtensions{.eps}     
\fi%

\graphicspath{{figures/}{pictures/}{images/}{./}} 

\usepackage{microtype}                 
\PassOptionsToPackage{warn}{textcomp}  
\usepackage{textcomp}                  
\usepackage{times}                     
\usepackage{cite}                      
\usepackage{tabu}                      
\usepackage{booktabs}                  

\usepackage{amsmath}
\usepackage{amssymb}
\usepackage{subfigure}
\usepackage[ruled,vlined,linesnumbered]{algorithm2e}
\usepackage{enumitem}
\usepackage{multicol}

\DeclareMathOperator*{\argmin}{argmin}

\onlineid{2181}

\vgtccategory{ISMAR 2021 Conference Papers (Technology)}

\vgtcinsertpkg

\title{Visual SLAM with Graph-Cut Optimized Multi-Plane Reconstruction}

\author{Fangwen Shu \thanks{e-mail: \{first\_name\}.\{last\_name\}@dfki.de}
\and
Yaxu Xie
\and
Jason Rambach
\and
Alain Pagani
\thanks{\textbf{Acknowledgment}: This research is partially funded by the German BMBF project MOVEON (01IS20077) and SocialWear (01IW20002).}
\and
Didier Stricker
}
\affiliation{\scriptsize DFKI - German Research Center for Artificial Intelligence}

\abstract{
This paper presents a semantic planar SLAM system that improves pose estimation and mapping using cues from an instance planar segmentation network. While the mainstream approaches are using RGB-D sensors, employing a monocular camera with such a system still faces challenges such as robust data association and precise geometric model fitting. In the majority of existing work, geometric model estimation problems such as homography estimation and piece-wise planar reconstruction (PPR) are usually solved by standard (greedy) RANSAC separately and sequentially. However, setting the inlier-outlier threshold is difficult in absence of information about the scene (i.e. the scale). In this work, we revisit these problems and argue that two mentioned geometric models (homographies/3D planes) can be solved by minimizing an energy function that exploits the spatial coherence, i.e. with graph-cut optimization, which also tackles the practical issue when the output of a trained CNN is inaccurate.
Moreover, we propose an adaptive parameter setting strategy based on our experiments, and report a comprehensive evaluation on various open-source datasets.
} 

\CCScatlist{
\CCScatTwelve{Computing methodologies}{Artificial intelligence}{ }{}
\CCScatTwelve{Computer vision}{Computer vision tasks}{Vision for robotics}{}
}

\nocopyrightspace

\begin{document}



\firstsection{Introduction}
\label{sec:introduction}
\maketitle

Semantic planar SLAM has gained much attention in the last decade, especially for virtual reality (VR) systems and augmented reality (AR) applications. Although there has been intensive research on this topic, most of the current methods still focus on RGB-D sensors \cite{hsiao2017keyframe, kaess2015simultaneous, salas2014dense, taguchi2013point} with plane primitives extracted from depth images. While monocular methods \cite{wang2020relative, yang2019monocular, rambach2019slamcraft} still face several challenges and difficulties, such as low-texture scenes, dynamic foregrounds, pure rotation of the camera, various baseline between frames, and scale drift, where plane primitives can only be extracted from limited 3D information obtained. Existing methods either build upon indirect SLAM \cite{mur2017orb} or direct SLAM \cite{engel2014lsd}, and both are subjected to the challenges mentioned before. 

\begin{figure}[!t]
  \centering
  \subfigure[Monocular (\textbf{fr3\_st\_tex\_far}).]{\includegraphics[width=0.43\linewidth]{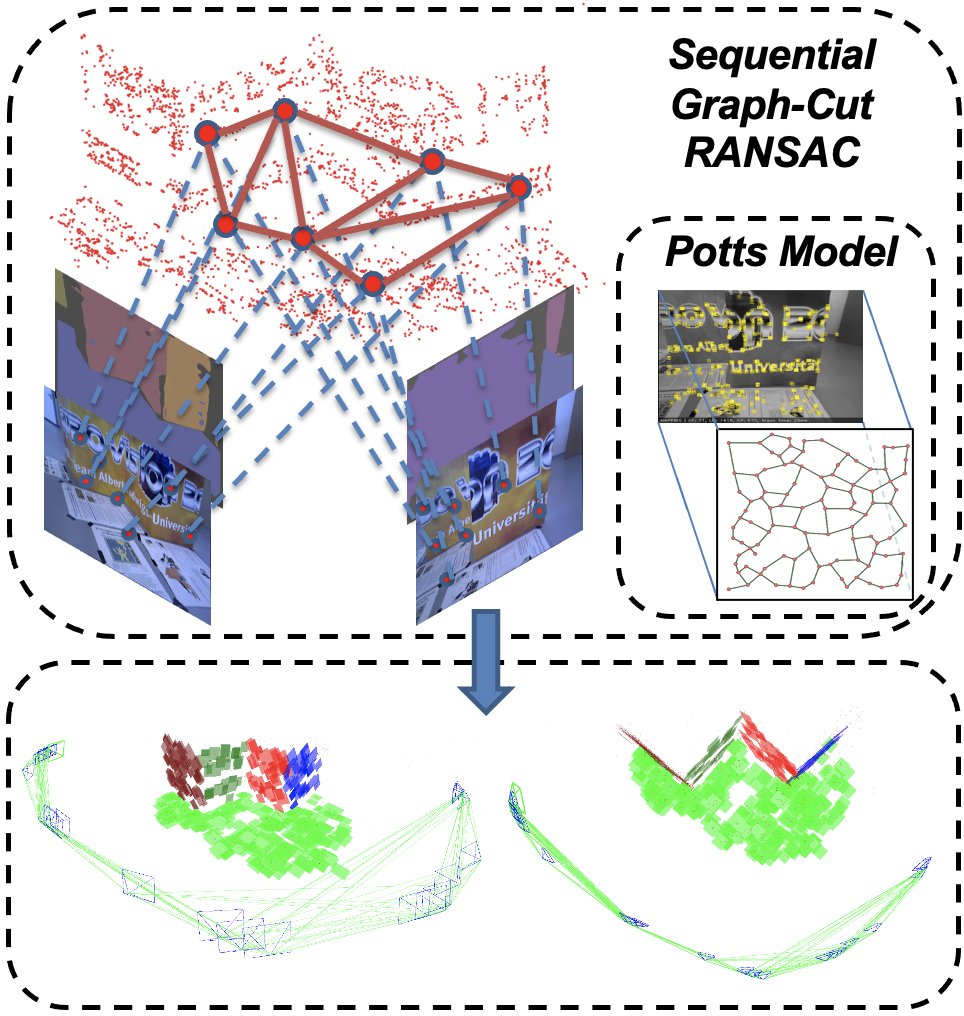}}
  \hfill
  \subfigure[RGB-D (\textbf{of\_kt3}).]{\includegraphics[width=0.49\linewidth]{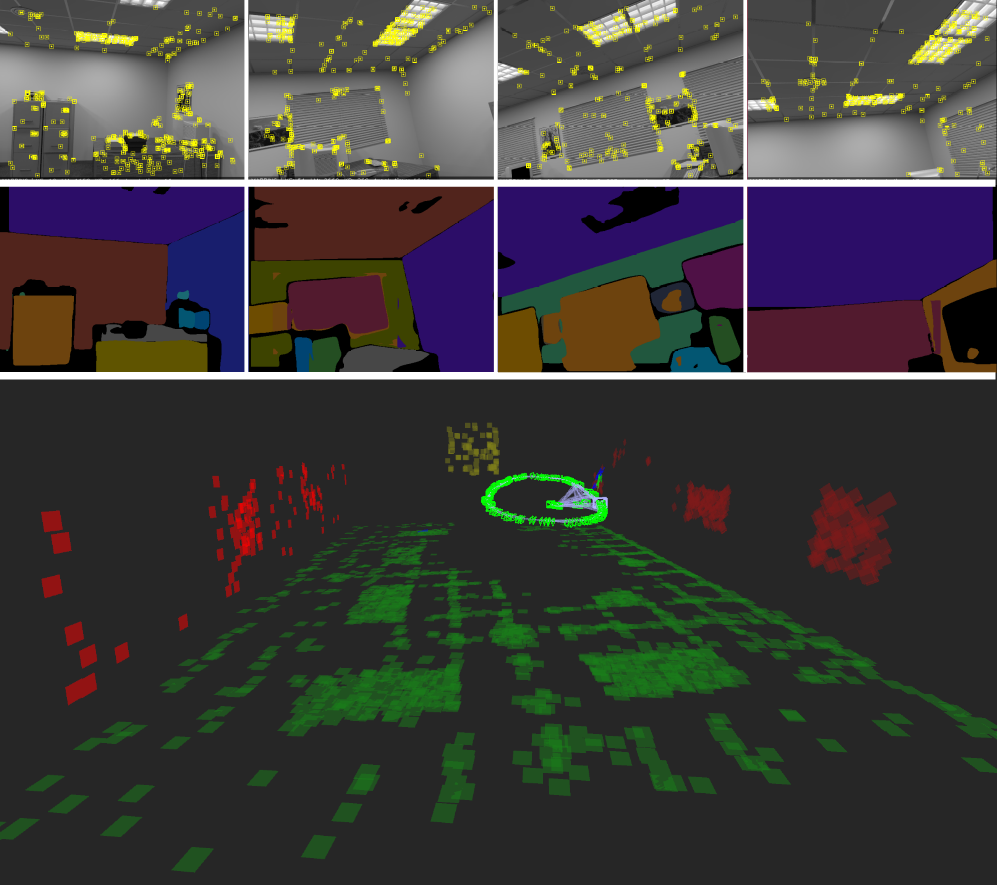}}
  \caption{\textbf{We propose using Sequential Graph-Cut RANSAC (Algo. \ref{alg:GC}) with feature-based SLAM for robust piece-wise planar reconstruction (PPR).} Here, the illustrated light-weight, planar, patch-surfels semantic map is reconstructed from sparse and noisy point clouds. Every distinguished color indicates a different plane.}
  \label{fig: overview}
\end{figure}

In this work, we argue that data association and geometric model fitting problems are usually not tackled efficiently in monocular SLAM systems, i.e. establishing feature matches of multi-plane between frames taken from different viewpoints (under small or large baseline), or from the same viewpoint (under pure rotation), with homographies estimated and decomposed. Thereafter, in order to localize the camera relatively, a plausible homography matrix is usually verified by triangulation (via positive depth validation) and minimizing symmetric transfer error (STE) between image pairs. However, the map scale is not observable purely from relative pose estimation. At the same time, 3D planes can only be fitted from sets of noisy and sparse point clouds triangulated by the monocular SLAM.
Thus, to tackle the problems especially for monocular systems, we first integrate a real-time instance planar segmentation network into a feature-based SLAM system. Then we propose to solve the multi-model fitting problem in a sequential RANSAC fashion but with a fast graph-cut optimized proposal engine \cite{barath2018graph}. 
Finally, we present quantitative and qualitative results from our semantic SLAM system, which shows the proposed method can be applied on any feature-based monocular or RGB-D SLAM system without a significant algorithm adaptation. To summarize, we propose the following contributions:

\begin{itemize}[leftmargin=*, noitemsep]
    \item We introduce an energy-based geometric model fitting method, i.e. sequential RANSAC with graph-cut optimization, into a feature-based planar SLAM system, which implicitly considers SLAM as optimizing geometric multi-model estimation of different types.
    \item We propose a SLAM building block that tightly integrates the energy-based method mentioned above and a state-of-the-art convolutional neural network (CNN) of instance planar segmentation. Thus, we do not take any output from CNN as a noise-free "sensor" measurement, but further optimize it within the SLAM workflow, which boosts the performance of tracking and optimization.
    \item We conduct exhaustive experiments and report a comprehensive evaluation on various indoor datasets.
\end{itemize}


\section{Related Work}
\label{sec:related}

\noindent \textbf{RGB-D Planar SLAM.}
There is a large body of recent and ongoing research on planar SLAM using 3D sensors. A common scenario of using semantic planar cues in SLAM is adding geometric regularization regarding different landmarks and optimize the geometric structure jointly, such as with Manhattan World (MW) assumption \cite{yunus2021manhattanslam}.
An early work~\cite{taguchi2013point} presented a SLAM system for a hand-held 3D sensor, where the authors argued that it is possible to register 3D data in two different coordinates using any combination of point and plane primitives.
This was followed by the works \cite{hsiao2017keyframe, kaess2015simultaneous, salas2014dense, zhang2019point} which tackle the problem similarly by extracting plane primitives from depth image, registering planes across different views, and optimize the poses of keyframes and landmarks (both point and plane) in Bundle Adjustment (BA).

\noindent \textbf{Monocular Planar SLAM.}
Even if the MW assumption is a good constraint for indoor SLAM, it is difficult to enforce it in monocular methods because only limited 3D information can be obtained. 
Therefore, this strong assumption is generally not used in the case of a monocular sensor. In this work, we focus more on developing a robust multi-plane estimator and refinement strategy.
For example, $\pi$Match \cite{raposo2016pi} employs PEARL (an energy-based geometric multi-model fitting method \cite{isack2012energy}) for piece-wise planar reconstruction in a novel two-view Structure-from-Motion (SfM) workflow. 
A recent monocular SLAM framework \cite{yang2019monocular} uses the high-level object and plane landmarks,
leading to a dense, compact, and semantically meaningful map compared to the classic feature-based SLAM. 
To track in a low-texture environment, Structure-SLAM \cite{li2020structure} employs not only planes but also predicted normals and lines to calculate drift-free rotation.
More recently, \cite{wang2020relative} proposed a Winner-Takes-All (WTA) RANSAC-based relative camera pose estimation under multiple planar structures with the help of superpixels segmentation. 

\noindent \textbf{Exploiting the Spatial Coherence in RANSAC.}
This kind of method is usually formalized as a binary labeling problem for geometric multi-model fitting, which we will discuss in detail in Sec. \ref{sec:implementation}. First with PEARL (Propose Expand and Re-estimate Labels) \cite{isack2012energy}, then with more advanced methods such as Graph-Cut RANSAC \cite{barath2018graph} and Progressive-X \cite{barath2019progressive}, the spatial coherence is exploited in the local optimization step to find local structures accurately, which consider that geometric data often form spatially coherent structures described by, e.g., the Potts model \cite{boykov2001fast}. Those methods are suitable for our problems in both 2D and 3D space, where homographies and piece-wise planes should not have spatial overlapping. 

\noindent \textbf{Plane Detection and Reconstruction via CNN.}
Initially PlaneNet \cite{liu2018planenet} presented a end-to-end neural architecture for piece-wise planar reconstruction from a single RGB image. However, it suffers from limitations such as missing small surfaces and requiring the maximum number of planes in a single image as prior. PlaneRCNN \cite{liu2019planercnn} addressed these issues and proposed the first detection-based neural network for piece-wise planar reconstruction, which jointly refines all the segmentation masks with warping loss function to enforce the consistency with a nearby view during training. It is able to detect small planar surfaces, but fails to reach a real-time frame rate. More recently, PlaneSegNet \cite{xie2021planesegnet} was proposed as a real-time one-stage instance plane segmentation network that achieves significantly higher frame rates and comparable segmentation accuracy against the two-stage methods mentioned before.

\section{Method}
\label{sec:implementation}

\begin{figure}[!t]
    \centering
    \includegraphics[width=0.95\linewidth]{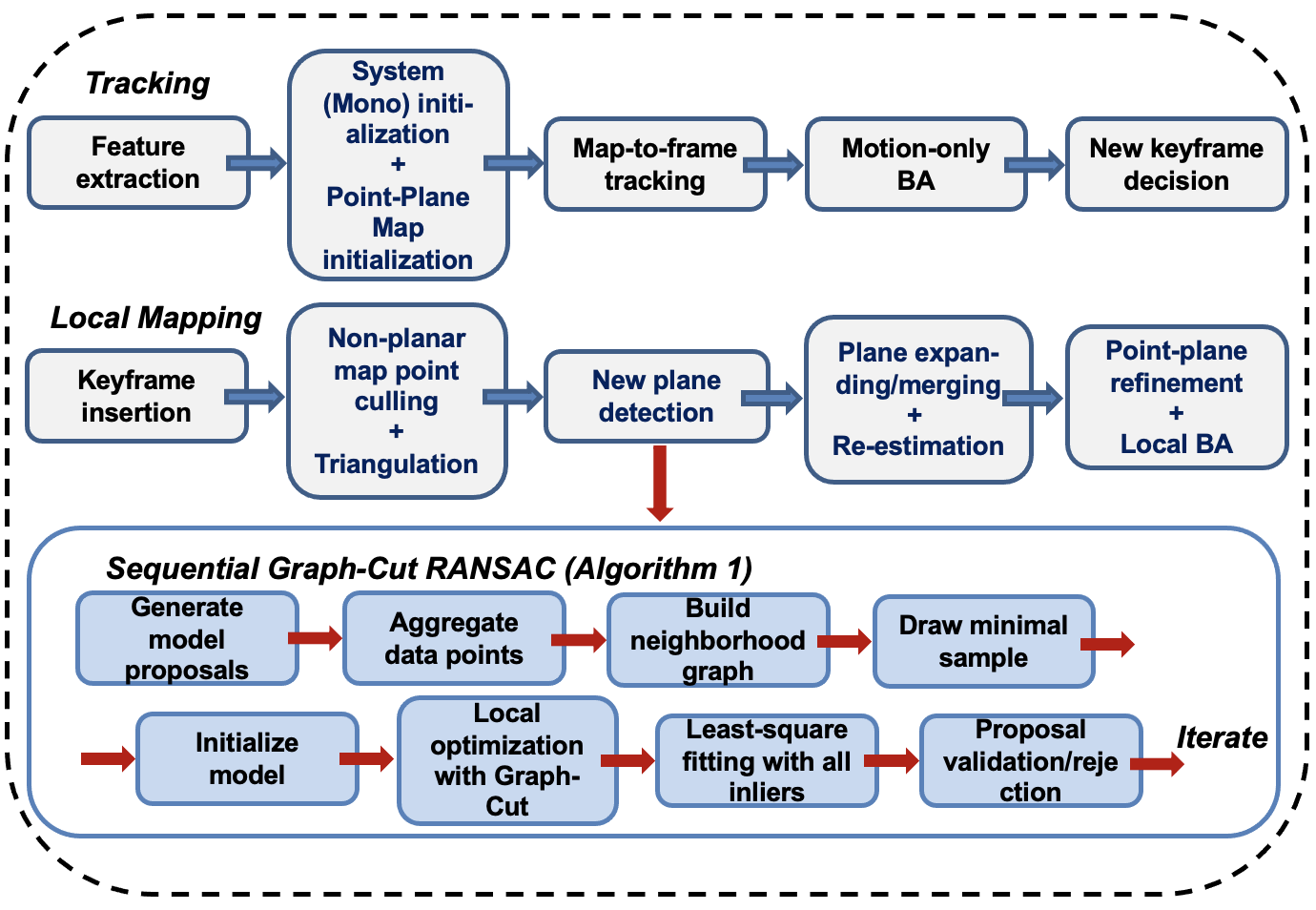}
    \caption{\textbf{The overall structure/workflow} of the monocular planar SLAM system used in this work.}
    \label{fig: BigDiagram}
\end{figure}

We first introduce the standard sequential RANSAC pipeline for geometric model fitting (homography or plane structure in this work), with semantic cues as input with image sequences. However, we would like to cope with possible misclassification from the instance segmentation network, and therefore we do not simply use a standard RANSAC-like plane fitting algorithm for each detected planar segment. Instead, we propose a more robust sequential pipeline using a locally optimized RANSAC alternating graph-cut and model re-fitting in the inner local optimization step (Algo. \ref{alg:GC}) to adapt automatically to inaccurate instance segmentation and noise. Finally, we discuss how it is integrated into a feature-based SLAM framework (as shown in Fig. \ref{fig: BigDiagram}) as a robust geometric multi-model fitting strategy in this work. The mathematical notation used in this section is adopted from \cite{barath2018graph, isack2012energy}.

\subsection{Geometric Model Fitting via RANSAC}
\label{sec: single-ransac}

Standard RANSAC \cite{fischler1981random} is a well-known method for dealing with a large number of outliers when data supports only one model (e.g. 2D line fitting). The implied unary energy: 
$E_{\{0;1\}}(L) = \sum_{p}^{} \|L_p\|_{\{0;1\}}$
counts inliers for the target model using 0-1 measure, thus can be reformulated as binary labeling problem \cite{isack2012energy}, where parameter vector $\theta$ of the model with the largest number of inliers within some threshold $\epsilon$ is estimated:
$$
	||L_p||_{\{0;1\}} = 
    \begin{cases}
		0 & \text{if } (L_p = 1 \wedge dist(p, \theta) < \epsilon) \text{ } \vee \\    
		 & \text{\phantom{xx}} (L_p = 0 \wedge dist(p, \theta) \geq \epsilon) \\    
        1 & \text{otherwise.}
	\end{cases}
$$
where $||L_p||_{\{0;1\}}$ is the geometric error measure, with $dist(p, \theta)$ the distance function, i.e. the Euclidean distance between data point $p \in P$ and the estimated model. Parameter $L \in \{0,1\}^{|P|}$ is the labeling, $|P|$ is the number of data points. $L_p \in L$ is the label of the point $p$. Here, the unary energy penalize nothing when $p$ is labeled as inlier (close to the model) or $p$ is labeled as outlier (far from the model).

\noindent \textbf{Sequential RANSAC.}
As a variant \cite{zuliani2005multiransac} of RANSAC, sequential RANSAC detects model instances one after another, with the inliers of the detected instance removed from data point set $P$. The drawback of this approach is that the inliers are usually assigned to the model instance which has the most support instead of the actual best instance. 
When we apply it with semantic priors, the proposed model’s inliers are already clustered from the data point set, thus it is followed by a simple validation of the model. The result is, obviously, dependent on the generalizability of a trained CNN and the user-defined thresholds that are difficult to fine-tune.

\subsection{Sequential Model Fitting with Spatial Coherence}
\label{sec: multi-ransac}

\begin{algorithm}[!t]
\small
\SetAlgoLined
\SetKwInOut{Input}{Input}
\SetKwInOut{Output}{Output}
\Input{$M_{seg}$ -- segmentation mask\; 
$P$ -- extracted feature points or matched correspondences\; 
$N$ -- max. iteration number of outer RANSAC loop\; 
$N_{GC}$ -- max. iteration number of inner GC-RANSAC loop\; 
$\epsilon_d$ -- point-to-model distance threshold\; 
$\epsilon_L$ -- residual threshold of a model\;}
\BlankLine
\Output{$\theta$ -- parameters of model instances\;}
\BlankLine
\For{$p \in P$}
{
    $L_p \leftarrow$ Generate model (labeling) proposals using $M_{seg}$\;
    $V_p \leftarrow$ Aggregate the data point\;
}
\For{$L_p \in \boldsymbol{L}$}
{
    $e_{L}^* \leftarrow$ Initialize residual with the max. numeric value\;
    $\mathcal{N} \leftarrow $ Build neighborhood-graph from $V_p$\;
    \For{$i = 1 \rightarrow$ $N$}
    {
        \For{$k = 1 \rightarrow$ $N_{GC}$}
        {
            $S_k \leftarrow$ Draw a minimal sample from $V_p$\;
            $\theta_k \leftarrow$ Estimate model parameters using $S_k$\;
            $w_k \leftarrow$ Find the inliers of $\theta_k$ using $\epsilon_d$\; 
            $e_{L_k} \leftarrow$ Calculate the residuals\;
            \If{$e_{L}^* > e_{L_k}$}
            {
                $e_{L}^*, \theta_{k}^*, w_{k}^* \leftarrow  e_{L_k}, \theta_k, w_k$\;
                \If{Local opt. (refer to Algo. 2 \cite{barath2018graph})}
                {
                    $w_{LO}^*, changed \leftarrow w_k, 1$\;
                    \While{$changed$}
                    {
                        $G \leftarrow $ Build the problem graph (refer to Algo. 3 \cite{barath2018graph})\;
                        $L_{LO} \leftarrow $ Apply graph-cut to $G$\;
                        $I_{7m} \leftarrow $ Select a $7m$-sized random inlier set\;
                        $\theta_{LO} \leftarrow $ Fit a model using $I_{7m}$\;
                        $w_{LO} \leftarrow $ Compute the support of $\theta_{LO}$\;
                        $changed \leftarrow 0$\;
                        \If{$w_{LO} > w_{LO}^*$}
                        {$\theta_{LO}^*, L_{LO}^*, w_{LO}^*, changed \leftarrow \theta_{LO}, L_{LO}, w_{LO}, 1$\;}
                    }
                    $L_{p}^*, \theta_{k}^*, w_{k}^* \leftarrow L_{LO}^*, \theta_{LO}^*, w_{LO}^*$\;
                }
            }
            $e_{L}^*, \theta^* \leftarrow $ Least squares fitting (SVD) using $w_{k}^*$\;
        }
        
        \If{$e_{L}^* < \epsilon_L$}
        {
            Early break;
        }
 }
}
\caption{\textbf{Sequential Graph-Cut RANSAC.}} 
\label{alg:GC}
\end{algorithm} 

The geometric multi-model fitting problem is usually formulated as an optimal labeling problem, where the binary energy $E(L)$ can be extended with an additional term indicating the label count penalty (label smoothness) \cite{li2007two} and a term indicating the spatial regularity \cite{isack2012energy}. The existing methods usually consider the general case when the data supports some unknown number of models, and solving $E(\boldsymbol{L})$ over labeling $\boldsymbol{L} = \{L_p | p \in P\}$ which describes the overall quality of the solution. However, in our case, the number of models is known from CNN, thus making the problem slightly different. 

\noindent \textbf{Energy Formulation.}
In this work, we assume the number of models is known, so there is no label count penalty added. We, therefore, formulate geometric fitting problems simply by optimizing the energy $E(L)$ over $K$ different models ($L_1, L_2, ...L_K$), in a sequential way.
The implicit assignment of inliers to models becomes trivial in our case because of an available segmentation prior, which means generating a large number of proposed labels (models) done in PEARL \cite{isack2012energy} is not needed anymore, making the algorithm very fast. 
In this case, for each model (labeling) proposal $L$, the assigned data point $p$ will be labeled as inlier or outlier after a so-far-the-best-model is found. As both inliers and outliers of the model should be spatially coherent, which means a point near an outlier (resp. inlier) is more likely to be an outlier (reps. inlier). In order to take this into account, we propose to use a graph-cut algorithm to further optimize the inlier-outlier assignment. The proposed energy:
\begin{equation}
    E(L) = \sum_{p}^{} \|L_p\| + \lambda \cdot \sum_{(p,q) \in \mathcal{N}}{w_{pq} \cdot \delta (L_p \neq L_q)}
    \label{equation: energy}
\end{equation}
where the first term indicates the geometric error measure between data point and corresponding model, and the second term indicates the spatial regularization which penalizes neighbors with different labels in the graph. $\mathcal{N}$ indicates edges in the near-neighbor graph constructed from data point set (e.g. the Potts model in Fig. \ref{fig: overview}). $\delta (\cdot)$ is 1 if the specified condition inside parenthesis holds, and 0 otherwise. Weights $w_{pq}$ set discontinuity penalties for each pair of neighboring data points. $\lambda$ is a parameter balancing the two terms.

The binary labeling energy minimization with additional spatial regularity terms (Eq. \ref{equation: energy}) can be solved efficiently and globally via the graph-cut algorithm. We adopted the idea of conducting the graph-cut algorithm in the local optimization (LO) step which is applied when a so-far-the-best model is found \cite{barath2018graph}. As the local optimization \cite{chum2003locally} assumes not all all-inlier samples are "good", it is perfectly aligned with our problem in this work, where the instance segmentation clusters the data samples in the first stage is prone to be inaccurate. More importantly, it is real-time feasible to apply the graph-cut in the LO step within just a few iterations, as the local optimization step converges very fast when it takes spatial proximity into account. 
The proposed \textit{Sequential Graph-Cut RANSAC} is presented in Algo. \ref{alg:GC}. The whole pipeline can be considered as several steps: (1) generate model proposals (labeling $L_1, L_2, ...L_k$) based on segmentation mask and assign the data points to the model; (2) estimate each model proposal sequentially, where the construction of problem graph $G$ (line 19) within the local optimization is used to build energy minimization of Eq. \ref{equation: energy}; (3) thereafter the graph-cut is applied to $G$ determining the optimal labeling $L$; (4) model parameters are updated according to, not only the number of the support inliers $w$ as done in \cite{barath2018graph}, but also a residual threshold $\epsilon_L$ defined within the SLAM system used in this work. Please note that the used parameters and thresholds will be further explained and detailed in the experiments (Sec. \ref{sec: result-slam}).

\subsection{Visual SLAM Framework}
\label{sec: vslam}

\noindent \textbf{System Initialization and Map Initialization.}
For monocular SLAM, we establish the proposed Algo. \ref{alg:GC} within the initialization step, where the Homography matrix and Fundamental matrix are calculated in parallel as done in \cite{mur2017orb}. 
We use symmetric transfer error (STE) for the geometric error measure $\|H_p\|$ between matched feature points $p = (p_{ref}, p_{cur})$. The initial solution for the non-linear minimization is found by using the Normalized Direct Linear Transform (NDLT) with the minimal 4 correspondences. Then we apply the energy minimization (Eq. \ref{equation: energy}) for homographies:
\begin{equation}
    E(\textbf{H}) = \sum_{p}^{} \|H_p\| + \lambda \cdot \sum_{(p,q) \in \mathcal{N}}{\delta (H_p \neq H_q)}
    \label{equation: H}
\end{equation}
where $\textbf{H} = \{ H|p \in P \}$ is the assignment of models to feature points $p$ in the reference frame, the neighborhood system $\mathcal{N}$ is based on a grid-neighborhood construction on image space and the minimum samples (4 correspondences) are sampled by progressive-NAPSAC sampler \cite{barath2019progressivenapsac}  within that image grid.

The homography with the most inliers is used to calculate the score $S_H$ \cite{mur2017orb} and initialize the map. The fundamental matrix is calculated using the default implementation.
Thereafter the initial map is scaled by setting the median of the inverse of the depth as 1 before tracking the next frame. After re-scaling the map, several planes can be fitted from the 3D point cloud. To this aim, we apply the energy minimization (Eq. \ref{equation: energy}) again within Algo. \ref{alg:GC} for piece-wise planar reconstruction:
\begin{equation}
    E(\boldsymbol{\Pi}) = \sum_{v}^{} \|\Pi_v\| + \lambda' \cdot \sum_{(u,v) \in \mathcal{N}'}{\delta (\Pi_u \neq \Pi_v)}
    \label{equation: PL}
\end{equation}
where $\boldsymbol{\Pi} = \{ \Pi | v \in V \}$ is the assignment of plane models to map point, and $V$ indicates the set of 3D vertices. Here, we use the distance between a 3D point and a plane: $dist(v, \Pi) = | \frac{\boldsymbol{n}^T \cdot \boldsymbol{v} + d}{\|\boldsymbol{n}\|} |$ as geometric error measure in the first term $\|\Pi_v\|$, with the plane represented as $(\boldsymbol{n}^T, d)^T$, where $\boldsymbol{n}$ is the plane normal and $d$ is the distance to the world origin. The neighborhood system $ \mathcal{N}'$ is constructed using Fast Approximate Nearest Neighbors algorithm \cite{muja2009fast} according to a predefined sphere radius $r$ as the 3D grid is unknown, and the minimum samples (3 points) are sampled uniformly.

\noindent \textbf{Local Mapping with Plane Expanded and Re-estimated.}
The existing feature-based method (i.e. ORB-SLAM2 \cite{mur2017orb}) focuses on utilizing as much point data as possible. When a new keyframe is inserted, however, we remove (local) map point which is assumed being associated with a plane but its distance $dist(v, \Pi)$ is bigger than a threshold $\epsilon_d$. This step is conducted before triangulation and local BA, thus will not influence the stability of the system. 
Right after the new point landmark is triangulated (still before local BA), we conduct Algo. \ref{alg:GC} for piece-wise planar reconstruction. However, detecting outliers of the geometric model is somewhat heuristic in this work. To avoid the presence of planes with weak support in terms of number of points, we consider planes with a low number of 3D points as "weak" planes and we only keep high-quality planes in the map. The weak planes can be later merged into other plane instances or removed if they cannot be expanded.

A map refiner is running in a loop within the local mapping thread, checking all the 3D plane instances and trying to merge the closed planes. Two planes are merged if the following two conditions are met, first they should have nearly parallel normals: $|cos(\theta)| = |\frac{\boldsymbol{n_i}\boldsymbol{n_j}}{\|\boldsymbol{n_i}\|\|\boldsymbol{n_j}\|}| > T_{\theta}$ ($0 < T_{\theta} < 1$)
and second, they should be geometrically close to each other: $|\frac{d_i}{\|d_i\|} - \frac{d_j}{\|d_j\|}| < T_{d}$.
The new plane equation is updated in a RANSAC loop with 60\% randomly sampled associated point landmark. After that, all associated point landmarks will be projected on the plane by minimizing the point-plane distance via: $\hat{\boldsymbol{v}} = \boldsymbol{v} - dist(\boldsymbol{v}, \Pi_v) \cdot \frac{\boldsymbol{n}}{\|\boldsymbol{n}\|}$.

\noindent \textbf{Optimization.} Different Bundle Adjustment (BA) are considered: (1) Motion-only BA for map-to-frame tracking. (2) Local BA on keyframe will be conducted after new plane detected, existing plane merged/expanded and re-estimated.
Notice for the local BA we treat the optimization of the structure and the motion in a de-coupled way. It means the reconstructed plane is used to structure the map first, thereafter the map is used in the local BA for joint optimization.
(3) Global BA only happens after loop closure. While a final refinement could be conducted with cost function \cite{lee2011mav}:
\begin{equation}
\footnotesize
    \argmin_{C_i, v_j, \Pi_k}\sum_{j}\left\{\sum_{i} ||p_{ij} - Q(C_i, v_j)|| + \sum_{k} dist(v_j, \Pi_k) \right\}
\end{equation}
where the first term indicates the standard reprojection error and $Q(.)$ is the camera projection function. The second term indicates the point-to-plane distance. The 6-DOF camera pose is represented as Lie Algebra $C \in \mathfrak{se}(3)$.
$i,j,k$ are the number of camera views, 3D points and planes, respectively.
Notice that the plane needs to be parameterized as minimal representation such as spherical coordinates: $\Pi = (\phi = arctan(\frac{n_y}{n_x}), \psi = arcsin(n_z), d)$, where $\phi$ and $\psi$ are the azimuth and elevation angles of the plane normal. 

However, under monocular setting, this cost function is subject to outlier planes (e.g. data sequences such as fr1\_desk/fr2\_desk, explained in Sec. \ref{sec: result-slam}) and can become a large scale non-linear optimization (e.g. any corridor scenario or large environment). In this work, for the fair comparison with other semantic SLAM methods, we report quantitative results (in Table \ref{tab:ATE}) without the final global optimization, and focus on the local mapping of PPR which shows the direct impact of the PPR on the accuracy of estimated trajectory.

\begin{table*}[!htb]
    \begin{minipage}{.65\linewidth}
      \centering
    \scalebox{0.58}{
        \begin{tabular}{|l||c|c|c|c||c|c|c|c|c|c|}
    \hline 
     Dataset \& Sequences & \multicolumn{4}{c||}{\textbf{Monocular}} & \multicolumn{5}{c|}{\textbf{RGB-D}}\\\hline
     \textbf{TUM RGB-D \cite{sturm2012benchmark}} & \textbf{Ours} & ORB-SLAM2 & Open- & Structure- & \textbf{Ours} & ORB-SLAM2 & Open- & Manhattan & SP-SLAM\\
     & & \cite{mur2017orb} & VSLAM\cite{sumikura2019openvslam}  & SLAM \cite{li2020structure} & & \cite{mur2017orb} & VSLAM\cite{sumikura2019openvslam} & SLAM \cite{yunus2021manhattanslam} & \cite{zhang2019point}\\
     \hline \hline
     fr1\_xyz   & \textbf{0.93}   & \underline{0.99} & 1.12 & - & 1.02 & 1.20 & 1.25 & \underline{1.00}  & \textbf{0.93} \\ 
     fr1\_floor & \textbf{1.72}   & 2.99 & \underline{1.91} & - & \textbf{1.64} & 2.50 & \underline{1.79} & -     & -    \\
     fr1\_desk  & 1.91   & \textbf{1.41} & \underline{1.73} & - & 1.81 & \underline{1.73} & 1.86 & 2.70  & \textbf{1.43} \\
     fr2\_xyz   & \textbf{0.24}     & 0.27   & \underline{0.26}  & - & 1.73 & \textbf{0.37} & 1.71 & \underline{0.80}  & -\\
     fr2\_desk  & 1.28     & \underline{1.14}   & \textbf{0.90}  & - & 7.66   & \textbf{0.85} & 7.85 & \underline{3.70}  & -\\ 
     fr3\_st\_tex\_far   & \textbf{0.84} & \underline{0.95}  & 1.09   & 1.40 & \underline{1.08} & 1.19 & 1.10 & 2.20 & \textbf{0.97}\\
     fr3\_st\_tex\_near  & \textbf{1.24} & 1.29  & \underline{1.28}   & 1.40 & \textbf{0.83} & 1.21 & 0.91 & 1.20 & \underline{0.84} \\
     fr3\_nst\_tex\_near & \underline{1.44} & \textbf{1.31}  & 2.50   & - & \textbf{1.02} & 2.25 & \underline{1.42} & -    & - \\
     \hline
     \hline
     \textbf{ICL-NUIM \cite{handa2014benchmark}} & \multicolumn{4}{c||}{} & \multicolumn{5}{c|}{}\\
     \hline \hline
     lr\_kt0 & \textbf{0.35} & \underline{0.37}   & \underline{0.37}  & -    & \textbf{0.46}  & 1.00 & 0.85 & \underline{0.70} & 0.80\\
     lr\_kt1 & 3.96  & \textbf{1.04}   & 2.40  & \underline{1.60} & \underline{0.72}  & 1.16 & \textbf{0.64} & 1.10 & 0.98\\
     lr\_kt2 & \textbf{2.62}  & 2.78   & \underline{2.72}  & 4.50 & \textbf{1.43} & 1.67 & 1.57 & \underline{1.50} & 1.92\\
     lr\_kt3 & \textbf{1.31}  & \underline{1.48}   & 2.35  & 4.60 & 1.58 & \textbf{1.02} & 1.77 & \underline{1.10} & 1.25\\
     of\_kt0 & \textbf{2.60}  & \underline{4.44}   & 6.70   & - & \textbf{1.92} & 2.54 & 2.19 & 2.50 & \underline{1.99}\\
     of\_kt1 & X  & X      & X      & X    & 3.15 & 5.64 & \underline{1.89} & \textbf{1.30} & 2.25\\
     of\_kt2 & 3.17  & \textbf{2.18}   & 4.54   & \underline{3.10} & 1.59 & \underline{0.97} & \textbf{0.87} & 1.50 & 2.20\\
     of\_kt3 & \underline{11.1}  & 18.03  & 13.50  & \textbf{6.50} & \textbf{0.91} & 6.94 & \underline{0.96} & 1.30 & 1.84\\
     \hline
\end{tabular}
}
\caption{\textbf{Absolute trajectory error (ATE) RMSE [cm]} (X stands for tracking failure, - stands result not available from the corresponding paper). Each result from ours, ORB-SLAM2 and OpenVSLAM was calculated as the average over 5 executions on each sequence.}
\label{tab:ATE}
    \end{minipage}%
    \hfill
    \begin{minipage}{.32\linewidth}
      \centering
    \scalebox{0.5}{
    \begin{tabular}{|l|c|c|c|}
    \hline 
        \textbf{Thread} & \textbf{Ours} & ORB-SLAM2 \cite{mur2017orb} & OpenVSLAM \cite{sumikura2019openvslam} \\
    \hline 
    \hline
        Tracking & \textbf{16.12} & 20.43 & 19.47 \\
        Local Mapping & \textbf{83.23} & 110.82 & 105.45 \\
    \hline
    \hline
    \textbf{Functionality (refer to Fig. \ref{fig: BigDiagram})} & \multicolumn{3}{c|}{\textbf{Ours}} \\
    \hline
    \hline
    Instance Planar Segmentation & \multicolumn{3}{c|}{33.11} \\
    System (Mono) Initialization & \multicolumn{3}{c|}{5.37} \\
    Point-Plane Map Initialization & \multicolumn{3}{c|}{22.14} \\
    Non-Planar Map Point Culling & \multicolumn{3}{c|}{0.56} \\
    New Plane Detection & \multicolumn{3}{c|}{2.07} \\
    Plane Merging/Expanding & \multicolumn{3}{c|}{0.92} \\
    Plane Re-estimation & \multicolumn{3}{c|}{1.11} \\
    Point-Plane Refinement & \multicolumn{3}{c|}{0.08} \\
    Local BA & \multicolumn{3}{c|}{59.85} \\
    \hline
    \end{tabular}
    }
    \caption{\textbf{Runtime analysis [ms] (mean value evaluated on dataset TUM RGB-D \cite{sturm2012benchmark}: fr3\_st\_tex\_far)} of our planar SLAM system compared to original ORB-SLAM2 and OpenVSLAM, \textbf{under monocular setting}, using a desktop PC with an Intel Xeon(R) E-2146G 12 cores CPU @ 3.50GHz, 32GB RAM. The PlaneSegNet is evaluated on a standard GPU of NVIDIA GTX 1080 Ti.}
    \label{tab: runtime}
    \end{minipage} 
\end{table*}

\begin{figure}[!t]
  \centering
  \includegraphics[width=0.75\linewidth]{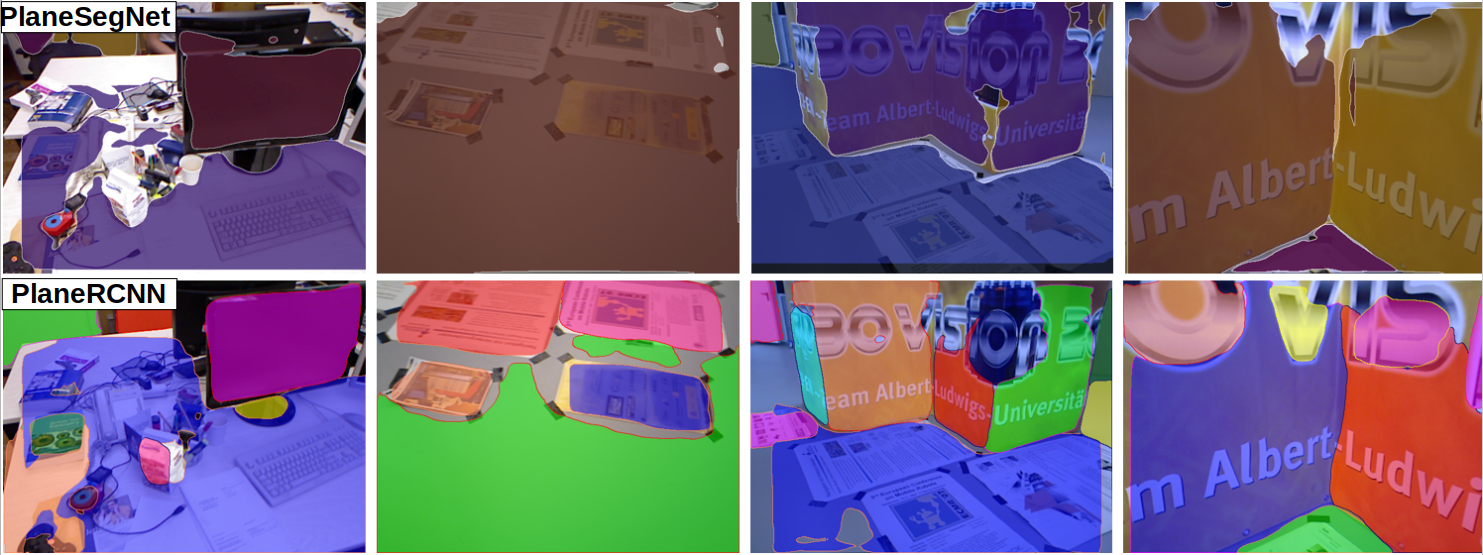}
  \caption{\textbf{Comparison of the segmentation results} of PlaneSegNet \cite{xie2021planesegnet} and PlaneRCNN \cite{liu2019planercnn} on dataset TUM RGB-D \cite{sturm2012benchmark}.}
\label{fig:psn_rcnn}
\end{figure}

\begin{figure}[!t]
  \centering
\subfigure[fr3\_st\_tex\_far.]{\includegraphics[width=0.37\linewidth]{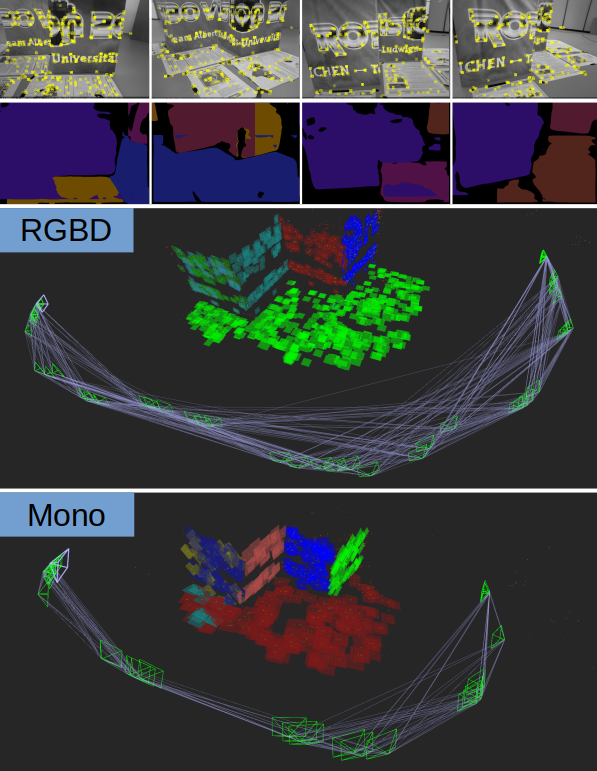}}
\subfigure[fr3\_st\_tex\_near.]{\includegraphics[width=0.39\linewidth]{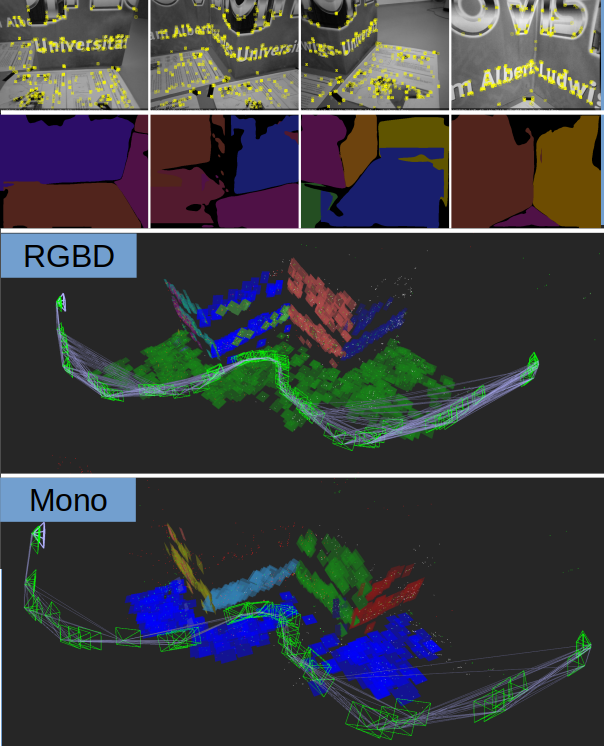}}
     
\subfigure[fr3\_nst\_tex\_near.]{\includegraphics[width=0.38\linewidth]{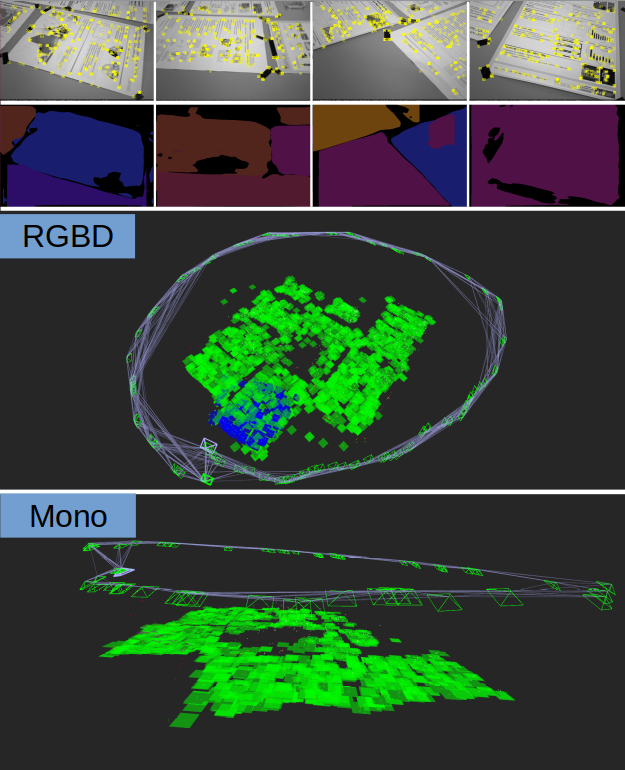}}
\subfigure[fr2\_xyz.]{\includegraphics[width=0.38\linewidth]{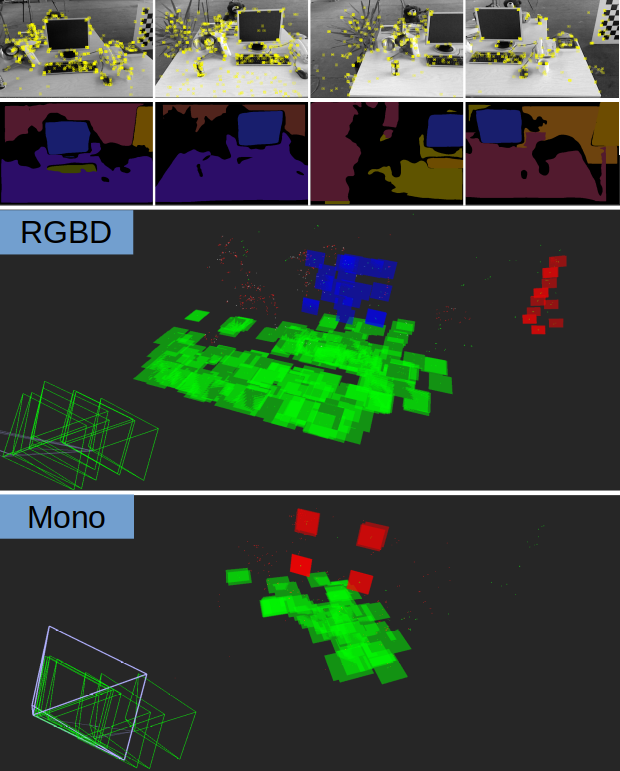}}
      
\subfigure[lr\_kt0.]{\includegraphics[width=0.4\linewidth]{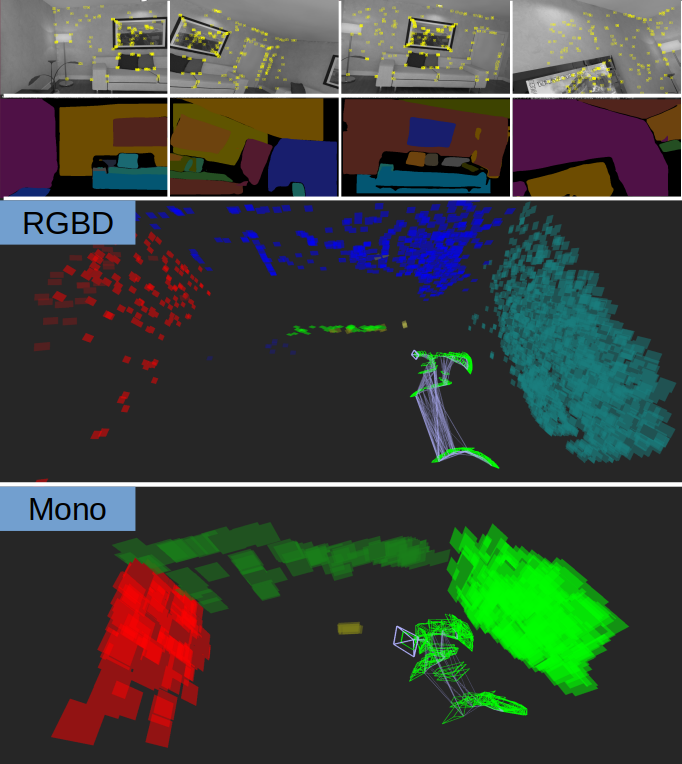}}
\subfigure[lr\_kt2.]{\includegraphics[width=0.36\linewidth]{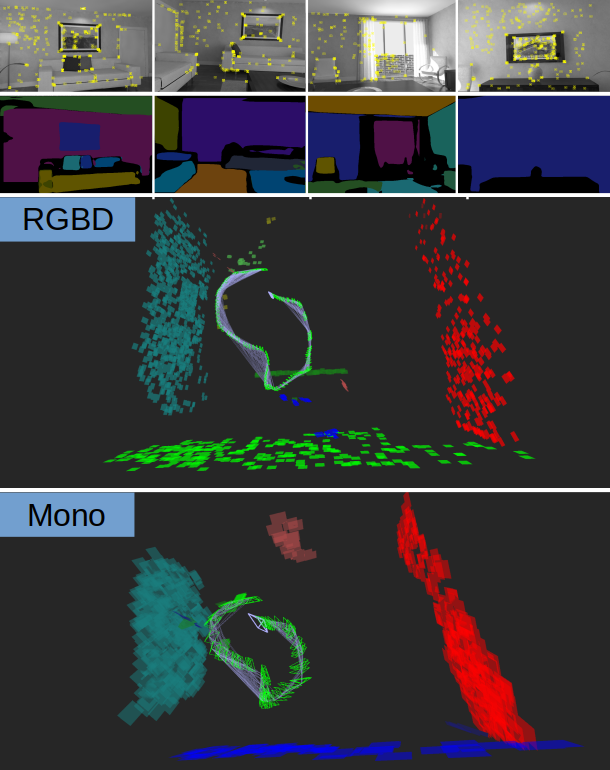}}
  \caption{\textbf{The light-weight semantic map (points and planar patch-surfels, best view zoom-in)} constructed on selected sequences of dataset TUM RGB-D \cite{sturm2012benchmark} and ICL-NUIM \cite{handa2014benchmark}.}
  \label{fig: map_tum}
\end{figure}


\section{Experiments and Results}
\label{sec:results}

We base our experiments on datasets TUM RGB-D \cite{sturm2012benchmark} and ICL-NUIM \cite{handa2014benchmark}. First, we discuss the chosen instance plane segmentation CNN, especially highlighting the problematic failure cases and explain how we treat it as noisy "sensor" measurement. Then we compare the performance of our SLAM framework with the proposed multi-plane estimator to some recent vSLAM methods. 

\subsection{Plane Detection via CNN}
\label{sec: result-cnn}

We employ PlaneSegNet~\cite{xie2021planesegnet} as our plane detector, because, as a global-mask-based instance segmentation method, it provides segmentation masks with higher resolution and better completeness than local-mask based method like PlaneRCNN~\cite{liu2019planercnn}. 
More important, PlaneSegNet is capable to run at a real-time frame rate (over 30Hz), while PlaneRCNN is only able to run at less than 5Hz with the same hardware (NVIDIA GTX 1080 Ti).

The instance segmentation is considered as a prior information for piece-wise planar reconstruction. However, as shown in Fig. \ref{fig:psn_rcnn}, global-mask based instance segmentation method (i.e. PlaneSegNet) suffers from feature leakage, and sometimes cannot distinguish different planes of similar texture.
Notice the network was not trained on the dataset we used to evaluate SLAM, which simulates the practical situation because a trained CNN may not generalize under the different real-world scenarios. This is also the reason why we introduced the graph-cut method in this work which can be considered as a post-processing step to the instance segmentation. Thus, we do not take any output from the CNN as the noise-free measurement of the plane, but further, optimize it within the SLAM framework. 
We argue that our proposed Sequential Graph-Cut RANSAC (Algo. \ref{alg:GC}) with plane expanded and re-estimated in SLAM (Sec. \ref{sec: vslam}) is an elegant way to reconstruct semantic structure in any monocular SLAM system. 
Within that, we actually established a procedure where first the potential hypothesis (data points) are separated based on segmentation prior and graph-structured optimization, thereafter the detected geometric primitives are merged based on their geometric attributes during the reconstruction of the scene, on the fly.

\subsection{Visual SLAM System with Point and Plane}
\label{sec: result-slam}

In this work, we made modification discussed in Sec. \ref{sec: vslam} to OpenVSLAM \cite{sumikura2019openvslam}
, which is built upon ORB-SLAM2 \cite{mur2017orb}.

\noindent \textbf{Geometric Model Fitting via Energy Minimization.}
For conducting the energy-based graph-structure optimization, a neighborhood-graph $\mathcal{N}$ has to be built according to the model estimated, where the data point is sampled from. For homography estimation, we define the number of cells along each axis as 8 which is used to divide the image into a grid where the neighborhood-graph is built from. The spatial coherence weight, i.e. the parameter $\lambda$ in Eq. \ref{equation: H}, is set as 0.975. The problem graph construction (line 19 in Algo. \ref{alg:GC}) within the local optimization (LO) step is used to add energy terms into the function (Eq. \ref{equation: H}), where the first energy term (geometric error measure) is added by replacing the standard 0-1 loss with a Gaussian kernel function, which makes the problem close to maximum likelihood estimation, refer to Eq. 1 of \cite{barath2018graph}. The second energy term (spatial regularization) is added by applying pair-wise energy on a modified Potts model, refer to Eq. 3 of \cite{barath2018graph}. The error threshold for symmetric transfer error (STE) is set as 2 pixels in this work. The confidence of the inner GC-RANSAC is set as 0.99. For piece-wise planar reconstruction, the neighborhood-graph is constructed via FLANN with the sphere radius set as $r = 2 \cdot \epsilon_d$ (where $\epsilon_d$ is the distance threshold discussed in the next paragraph). The confidence of the inner GC-RANSAC is set as 0.99. The spatial coherence weight ($\lambda'$) is set as 0.6. For all the experiments, the max. iteration number $N_{GC}$ is calculated according to the Eq. 4 in \cite{barath2018graph}, the max. iteration number $N$ of outer RANSAC loop in Algo. \ref{alg:GC} is set as 50.

\noindent \textbf{Adaptive Parameter Setting Strategy.}
Setting the inlier-outlier threshold is difficult without knowing the scale. Thus, to reduce the dependency on the user-defined threshold, we set two empirical values but adjust it according to the local map scale dynamically: (1) the distance threshold $\epsilon_d$ which decides if a point landmark is close enough to a plane, (2) the residual error $\epsilon_{\Pi}$ which decides if a plane equation fitted is optimal enough (here, $\epsilon_{\Pi}$ is equivalent to the $\epsilon_{L}$ in Algo. \ref{alg:GC} which is the residual of a model). 
Given $\epsilon_d = 0.02$ and $\epsilon_{\Pi} = 0.01$, for monocular mode, the local map scale is estimated as the inverse of median depth of current keyframe during tracking, thereafter both thresholds $\epsilon_d$ and $\epsilon_{\Pi}$ are normalized by the local map scale on the fly. For RGB-D mode, the local map scale is calculated as the average value of the sum of the world position of all the point landmarks, this step only needs to be conducted at the beginning of the data sequence. Moreover, the geometric thresholds mentioned in the local mapping thread (Sec. \ref{sec: vslam}): $T_{\theta}$ = 0.8 (decides if two normals are parallel enough) and $T_d = 10 \cdot \epsilon_d$ (decides if two planes are close enough). We found out above mentioned thresholds give our SLAM system stable performance on various data sequences during evaluation, without much effort on parameter fine-tuning.

\noindent \textbf{Benchmarking.}
The TUM RGB-D benchmark \cite{sturm2012benchmark} provides indoor sequences under different texture and structure conditions. Thus, we select different levels of complexity for evaluating our planar SLAM system: single plane scenario (fr1\_floor, fr3\_nst\_tex\_near), multiple planes scenario (fr3\_st\_tex\_far, fr3\_st\_tex\_near), and scenario of textureless table but with many objects presented (fr1\_xyz, fr1\_desk, fr2\_xyz, fr2\_desk), as listed in Table \ref{tab:ATE}. 
While our monocular planar SLAM system obtains better results of ATE RMSE on most of the sequence compared to the classic feature-based SLAM systems. Notice that most of the relative work is not superior in terms of ATE RMSE compared to e.g. ORB-SLAM2, such as Structure-SLAM (Mono) and ManhattanSLAM (RGB-D) which reported less complete quantitative results in their paper. 
Nevertheless, Structure-SLAM, ManhattanSLAM, and SP-SLAM utilize not only plane features but also lines, predicted normal maps or MW assumption in their system. Ours, however, presented as a pure point-plane SLAM system without WM assumption whose algorithm is suitable for any feature-based SLAM of monocular or RGB-D, which shows integrating MW assumption or more high-level features (e.g. line) benefit semantic mapping or tracking in the low-texture scenes, but not necessarily improve the accuracy of camera localization, possibly due to different optimization strategies.  More important, any output from CNN should not be considered as noise-free input for the semantic SLAM system, as we especially tackled in this work with graph-cut optimization within SLAM workflow. The qualitative results are illustrated in Fig. \ref{fig: map_tum}.

The ICL-NUIM RGB-D benchmark \cite{handa2014benchmark} is a synthetic indoor dataset that shows a low-contrast and low-texture environment. Thus we lower the FAST threshold for detecting ORB features to 2, which gives the best performance under monocular setting,
while RGB-D SLAM is evaluated under the default setting. However, even with a lower FAST threshold, this dataset is difficult for monocular SLAM and our monocular system only works stably on a few of the sequences, and we are not able to initialize the system when processing more than half of the images of sequence of\_kt1. Without establishing 3D plane-plane registration and pose optimization \cite{taguchi2013point} as done in mainstream approaches of RGB-D planar SLAM, our RGB-D system also obtains comparable ATE RMSE accuracy compare to other RGB-D SLAM methods, as shown in Table \ref{tab:ATE}.

\noindent \textbf{Failure Cases and Limitation.}
Our monocular planar SLAM system depends strongly on the point features, which brings the limitation that no reliable plane can be fitted when there are not enough point landmarks. Our map refiner strategy only keeps high-quality planes, which also omits small planes from the map. Textureless planar scenes such as fr1\_desk and fr2\_desk, result in fitted planes from point cloud actually associated with objects like books, keyboards, and cups, which undermines the camera localization. The ATE RMSE evaluation of our SLAM systems also strongly depends on the performance of OpenVSLAM. The tracking failure of sequence of\_kt1 (monocular) results from a fast-moving and fast-rotating camera and textureless scene. RGB-D SLAM is usually more robust but more 3D point and plane landmarks could result in a large-scale non-linear optimization even in a small scene. 

\noindent \textbf{Run-time.}
The run-time analysis of our monocular SLAM system is reported in Table \ref{tab: runtime}, which shows a more efficient computation time compared to the classic feature-based ORB-SLAM2 and OpenVSLAM. Notice the functionality presented in Table \ref{tab: runtime} is corresponding to the building block illustrated in Fig. \ref{fig: BigDiagram}, which is embedded as a multi-thread system. Most of the computation is used for initialization and local BA. After the system is initialized, the instance planar segmentation only needs to be conducted on every inserted keyframe. It is very fast to detect new planes using our proposed sequential multi-plane fitting method (avg. 2 ms) when a new keyframe is inserted into the map, and it is very fast to merge, expand, and refine the existing plane structures (avg. 1 ms).

\section{Conclusion}
\label{sec:conclusion}

Our work presented a robust building block of a feature-based SLAM framework with an extended plane detector, with special care in taking instance plane segmentation as noisy "sensor" input and further optimizing it during geometric primitives reconstruction. With the dynamically adjusted thresholds, our proposed multi-plane reconstructor can be applied to various indoor scenarios without much effort in parameter fine-tuning. Comprehensive quantitative results are reported in this work. Future works can explore other types of features such as line segment and vanish point, or utilize planar SLAM in the urban environment and driving scenario. The joint pose-graph optimization of different geometric primitives is of interest.



\bibliographystyle{abbrv-doi}

\bibliography{template}

\begin{thebibliography}{10}

\bibitem{barath2019progressivenapsac}
D.~Barath, M.~Ivashechkin, and J.~Matas.
\newblock Progressive napsac: sampling from gradually growing neighborhoods.
\newblock {\em arXiv preprint arXiv:1906.02295}, 2019.

\bibitem{barath2018graph}
D.~Barath and J.~Matas.
\newblock Graph-cut ransac.
\newblock In {\em Proceedings of the IEEE conference on computer vision and
  pattern recognition}, 2018.

\bibitem{barath2019progressive}
D.~Barath and J.~Matas.
\newblock Progressive-x: Efficient, anytime, multi-model fitting algorithm.
\newblock In {\em Proceedings of the IEEE/CVF International Conference on
  Computer Vision}, pp. 3780--3788, 2019.

\bibitem{boykov2001fast}
Y.~Boykov, O.~Veksler, and R.~Zabih.
\newblock Fast approximate energy minimization via graph cuts.
\newblock {\em IEEE Transactions on pattern analysis and machine intelligence},
  23(11):1222--1239, 2001.

\bibitem{chum2003locally}
O.~Chum, J.~Matas, and J.~Kittler.
\newblock Locally optimized ransac.
\newblock In {\em Joint Pattern Recognition Symposium}, pp. 236--243. Springer,
  2003.

\bibitem{engel2014lsd}
J.~Engel, T.~Sch{\"o}ps, and D.~Cremers.
\newblock Lsd-slam: Large-scale direct monocular slam.
\newblock In {\em European conference on computer vision}, pp. 834--849.
  Springer, 2014.

\bibitem{fischler1981random}
M.~A. Fischler and R.~C. Bolles.
\newblock Random sample consensus: a paradigm for model fitting with
  applications to image analysis and automated cartography.
\newblock {\em Communications of the ACM}, 24(6):381--395, 1981.

\bibitem{handa2014benchmark}
A.~Handa, T.~Whelan, J.~McDonald, and A.~J. Davison.
\newblock A benchmark for rgb-d visual odometry, 3d reconstruction and slam.
\newblock In {\em 2014 IEEE international conference on Robotics and automation
  (ICRA)}, pp. 1524--1531. IEEE, 2014.

\bibitem{hsiao2017keyframe}
M.~Hsiao, E.~Westman, G.~Zhang, and M.~Kaess.
\newblock Keyframe-based dense planar slam.
\newblock In {\em 2017 IEEE International Conference on Robotics and Automation
  (ICRA)}, pp. 5110--5117. IEEE, 2017.

\bibitem{isack2012energy}
H.~Isack and Y.~Boykov.
\newblock Energy-based geometric multi-model fitting.
\newblock {\em International journal of computer vision}, 97(2):123--147, 2012.

\bibitem{kaess2015simultaneous}
M.~Kaess.
\newblock Simultaneous localization and mapping with infinite planes.
\newblock In {\em 2015 IEEE International Conference on Robotics and Automation
  (ICRA)}, pp. 4605--4611. IEEE, 2015.

\bibitem{lee2011mav}
G.~H. Lee, F.~Fraundorfer, and M.~Pollefeys.
\newblock Mav visual slam with plane constraint.
\newblock In {\em 2011 IEEE International Conference on Robotics and
  Automation}, pp. 3139--3144. IEEE, 2011.

\bibitem{li2007two}
H.~Li.
\newblock Two-view motion segmentation from linear programming relaxation.
\newblock In {\em 2007 IEEE conference on computer vision and pattern
  recognition}, pp. 1--8. IEEE, 2007.

\bibitem{li2020structure}
Y.~Li, N.~Brasch, Y.~Wang, N.~Navab, and F.~Tombari.
\newblock Structure-slam: Low-drift monocular slam in indoor environments.
\newblock {\em IEEE Robotics and Automation Letters}, 5(4):6583--6590, 2020.

\bibitem{liu2019planercnn}
C.~Liu, K.~Kim, J.~Gu, Y.~Furukawa, and J.~Kautz.
\newblock Planercnn: 3d plane detection and reconstruction from a single image.
\newblock In {\em Proceedings of the IEEE Conference on Computer Vision and
  Pattern Recognition}, pp. 4450--4459, 2019.

\bibitem{liu2018planenet}
C.~Liu, J.~Yang, D.~Ceylan, E.~Yumer, and Y.~Furukawa.
\newblock Planenet: Piece-wise planar reconstruction from a single rgb image.
\newblock In {\em Proceedings of the IEEE Conference on Computer Vision and
  Pattern Recognition}, pp. 2579--2588, 2018.

\bibitem{muja2009fast}
M.~Muja and D.~G. Lowe.
\newblock Fast approximate nearest neighbors with automatic algorithm
  configuration.
\newblock {\em VISAPP (1)}, 2(331-340):2, 2009.

\bibitem{mur2017orb}
R.~Mur-Artal and J.~D. Tard{\'o}s.
\newblock Orb-slam2: An open-source slam system for monocular, stereo, and
  rgb-d cameras.
\newblock {\em IEEE Transactions on Robotics}, 33(5):1255--1262, 2017.

\bibitem{rambach2019slamcraft}
J.~Rambach, P.~Lesur, A.~Pagani, and D.~Stricker.
\newblock Slamcraft: Dense planar rgb monocular slam.
\newblock In {\em 2019 16th International Conference on Machine Vision
  Applications (MVA)}, pp. 1--6. IEEE, 2019.

\bibitem{raposo2016pi}
C.~Raposo and J.~P. Barreto.
\newblock $\pi$match: Monocular vslam and piecewise planar reconstruction using
  fast plane correspondences.
\newblock In {\em European Conference on Computer Vision}, pp. 380--395.
  Springer, 2016.

\bibitem{salas2014dense}
R.~F. Salas-Moreno, B.~Glocken, P.~H. Kelly, and A.~J. Davison.
\newblock Dense planar slam.
\newblock In {\em 2014 IEEE international symposium on mixed and augmented
  reality (ISMAR)}, pp. 157--164. IEEE, 2014.

\bibitem{sturm2012benchmark}
J.~Sturm, N.~Engelhard, F.~Endres, W.~Burgard, and D.~Cremers.
\newblock A benchmark for the evaluation of rgb-d slam systems.
\newblock In {\em 2012 IEEE/RSJ International Conference on Intelligent Robots
  and Systems}, pp. 573--580. IEEE, 2012.

\bibitem{sumikura2019openvslam}
S.~Sumikura, M.~Shibuya, and K.~Sakurada.
\newblock Openvslam: A versatile visual slam framework.
\newblock In {\em Proceedings of the 27th ACM International Conference on
  Multimedia}, pp. 2292--2295, 2019.

\bibitem{taguchi2013point}
Y.~Taguchi, Y.-D. Jian, S.~Ramalingam, and C.~Feng.
\newblock Point-plane slam for hand-held 3d sensors.
\newblock In {\em 2013 IEEE international conference on robotics and
  automation}, pp. 5182--5189. IEEE, 2013.

\bibitem{wang2020relative}
X.~Wang, M.~Christie, and E.~Marchand.
\newblock Relative pose estimation and planar reconstruction via
  superpixel-driven multiple homographies.
\newblock In {\em IEEE/RSJ Int. Conf. on Intelligent Robots and Systems,
  IROS'20}, 2020.

\bibitem{xie2021planesegnet}
Y.~Xie, J.~Rambach, F.~Shu, and D.~Stricker.
\newblock Planesegnet: Fast and robust plane estimation using a single-stage
  instance segmentation cnn.
\newblock {\em arXiv preprint arXiv:2103.15428}, 2021.

\bibitem{yang2019monocular}
S.~Yang and S.~Scherer.
\newblock Monocular object and plane slam in structured environments.
\newblock {\em IEEE Robotics and Automation Letters}, 4(4), 2019.

\bibitem{yunus2021manhattanslam}
R.~Yunus, Y.~Li, and F.~Tombari.
\newblock Manhattanslam: Robust planar tracking and mapping leveraging mixture
  of manhattan frames.
\newblock {\em arXiv preprint arXiv:2103.15068}, 2021.

\bibitem{zhang2019point}
X.~Zhang, W.~Wang, X.~Qi, Z.~Liao, and R.~Wei.
\newblock Point-plane slam using supposed planes for indoor environments.
\newblock {\em Sensors}, 19(17):3795, 2019.

\bibitem{zuliani2005multiransac}
M.~Zuliani, C.~S. Kenney, and B.~Manjunath.
\newblock The multiransac algorithm and its application to detect planar
  homographies.
\newblock In {\em IEEE International Conference on Image Processing 2005}.
  IEEE, 2005.

\end{thebibliography}
\end{document}